\documentclass[letterpaper, 10 pt, conference]{formatting/ieeeconf}
\IEEEoverridecommandlockouts                              
\let\labelindent\relax
\usepackage{amsfonts}       

\usepackage{amsthm}
\usepackage{mathtools}      
\usepackage{amssymb}        
\usepackage{filecontents}
\usepackage{graphicx}       
\usepackage{marginnote}     
\usepackage{marvosym}       
\usepackage{overpic}        
\usepackage{tabularx}
\usepackage{cite}
\usepackage{color}
\usepackage[linesnumbered,algoruled,boxed,lined]{algorithm2e}
\usepackage[normalem]{ulem}
\usepackage{epstopdf}
\usepackage{enumitem}
\usepackage[normalem]{ulem}
\usepackage{lipsum}
\usepackage[a-2b,mathxmp]{pdfx}[2018/12/22]
\usepackage{comment}
\usepackage[font={footnotesize}]{caption}
\usepackage{subcaption}

\newif\ifdraft
\draftfalse

\ifdraft
\usepackage[paperheight=11in,paperwidth=9.5in,
			left=1.25in,right=1.25in,
			top=0.75in,bottom=0.75in,
			heightrounded,marginparwidth=1.2in,
			marginparsep=0.05in]{geometry}
\usepackage{xcolor}
\usepackage{xargs} 
\usepackage[textsize=footnotesize]{todonotes}
\newcommandx{\dz}[2][1=]{\todo[linecolor=red,
			backgroundcolor=red!10,bordercolor=red,#1]{DZ:#2}}
\newcommandx{\jy}[2][1=]{\todo[linecolor=green,
			backgroundcolor=green!10,bordercolor=green,#1]{JY:#2}}
\newcommandx{\jh}[2][1=]{\todo[linecolor=blue,
            backgroundcolor=blue!10,bordercolor=blue,#1]{JH:#2}}
\else
\newcommand{\dz}[1]{{}}
\newcommand{\jy}[1]{{}}
\newcommand{\jh}[1]{{}}
\fi

\newcommand*{\rom}[1]{\expandafter\@slowromancap\romannumeral #1@}

\newcommand{\set}[1]{\left\{ #1 \right\}}

\def\real{\mathbb{R}}

\newif\iftwocolumn
\twocolumntrue

\newtheorem{problem}{Problem}

\theoremstyle{definition}

\theoremstyle{remark}

\SetKwProg{Fn}{Function}{}{}
\SetKwComment{Comment}{$\triangleright$\ }{}

\makeatletter
\def\subsubsection{\@startsection{subsubsection}
                                 {3}
                                 {\z@ \hspace*{1mm}}
                                 {0ex plus 0.1ex minus 0.1ex}
                                 {0ex}
                                 {\normalfont\normalsize\itshape}}
\makeatother

\font\titlefont=ptmb at 13.9pt
\title{\titlefont
High-Performance Dual-Arm Task and Motion Planning for Tabletop Rearrangement
}
\author{
Duo Zhang \qquad Junshan Huang \qquad Jingjin Yu
\thanks{
D. Zhang, J. Huang, and J. Yu are with the Department of Computer Science, 
Rutgers, the State University of New Jersey, Piscataway, NJ, 
USA. E-Mails: \{{\tt duo.zhang, junshan.huang, jingjin.yu}\}\hspace*{.25em}
\MVAt \hspace*{.25em}rutgers.edu. This work is supported in part by AFOSR award FA9550-25-1-0249, NSF awards IIS-1845888, IIS-2132972, CCF-2309866, an Amazon Research award, and an NVIDIA Academic Grant.}
}

\def\ours{\textsc{SDAR}\xspace}
\def\ourst{\textsc{SDAR-T}\xspace}
\def\oursm{\textsc{SDAR-M}\xspace}

\begin{document}

\maketitle
\thispagestyle{empty}
\pagestyle{empty}


\begin{abstract}
We propose Synchronous Dual-Arm Rearrangement Planner (\ours), a task and motion planning (TAMP) framework for tabletop rearrangement, where two robot arms equipped with 2-finger grippers must work together in close proximity to rearrange objects whose start and goal configurations are strongly entangled. 
To tackle such challenges, \ours tightly knit together its dependency-driven task planner (\ourst) and synchronous dual-arm motion planner (\oursm), to intelligently sift through a large number of possible task and motion plans. 
Specifically, \ourst applies a simple yet effective strategy to decompose the global object dependency graph induced by the rearrangement task, to produce more optimal dual-arm task plans than solutions derived from optimal task plans for a single arm. 
Leveraging state-of-the-art GPU SIMD-based motion planning tools, \oursm employs a layered motion planning strategy to sift through many task plans for the best synchronous dual-arm motion plan while ensuring high levels of success rate. 
Comprehensive evaluation demonstrates that \ours delivers a $100\%$ success rate in solving complex, non-monotone, long-horizon tabletop rearrangement tasks with solution quality far exceeding the previous state-of-the-art. Experiments on two UR-5e arms further confirm \ours directly and reliably transfers to robot hardware. Source code and supplementary materials are available at \href{https://github.com/arc-l/dual-arm}{\texttt{\textcolor{blue}{https://github.com/arc-l/dual-arm}}}. 

\end{abstract}

\section{Introduction}\label{sec:intro}
Task and motion planning (TAMP) \cite{garrett2021integrated} represents a fundamental computation challenge in robotics, in which a robot system, e.g., one or more robot arms, must break down a given, potentially long-horizon task into suitable ``bite-sized'' sub-tasks that can be executed through short-horizon robot motions. 
TAMP, which integrates high-level reasoning with low-level motion control, stands as a cornerstone in robotics as it holds the promise to empower robots to carry out the full spectrum of daily human tasks that demand physical interaction with the world.
Yet, ``solving'' TAMP has remained elusive because doing so must handle the combinatorial explosion of discrete sub-task partitioning and sequencing, and simultaneously tackle the complexity of generating high-quality motion in high-dimensional environments, both of which are hard computational challenges themselves~\cite{han2018complexity,GaoFenHuaYu23IJRR,hopcroft1984complexity}.

Despite intractability obstacles, TAMP continues to attract significant research attention with remarkable progress being made  \cite{garrett2020pddlstream,gao2023orla,curtis2022long,zhang2022visual,wang2022hierarchical,shridhar2023perceiver,yang2025guiding,xu2025set}, which have explored solution schemes ranging from combinatorial search to data-driven, and anywhere in between. With an effective planning framework~\cite{garrett2020pddlstream}, speedy optimal task planners~\cite{WanGaoNakYuBek21ICRA,GaoLauHuaBekYu22ICRA}, and recent advances in CPU/GPU-SIMD accelerated motion planning~\cite{thomason2023motions,curobo}, the computation bottleneck for systems with a single robot arm has largely shifted from integrating task and motion planning to complex task/scene understanding.
\begin{figure}[t]
\centering
\begin{overpic}[width=\linewidth]{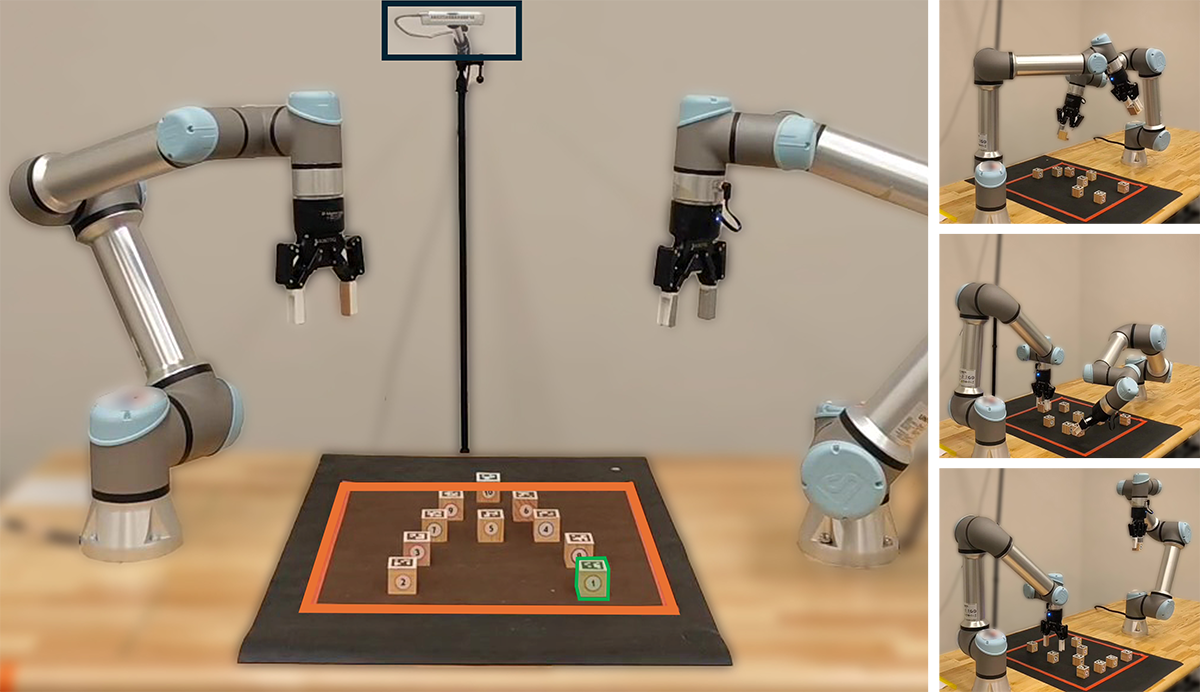}
  \put(45,54){\textcolor{yellow!90!black}{\bfseries Camera}}        
  \put(32,18){\textcolor{orange!100!black}{\bfseries Workspace}}     
  \put(52,8){\textcolor{green!70!black}{\bfseries Object}}         
  \put(6,50){\bfseries Arm1}                                        
  \put(62,50){\bfseries Arm2}                                       

  \put(78.5,55){\small (a)} 
  \put(78.5,35.2){\small (b)} 
  \put(78.5,15.9){\small (c)} 

\end{overpic}
\caption{Illustration of the dual-arm tabletop rearrangement setup examined in this work. 
%
(a) Motion planner performing regular pick-and-place, (b) Arms working in close proximity to handle objects that are close to each other, (c) Falling back to single-arm sequential execution when no feasible dual-arm solution can be quickly found.}
\label{fig:example}
\end{figure}
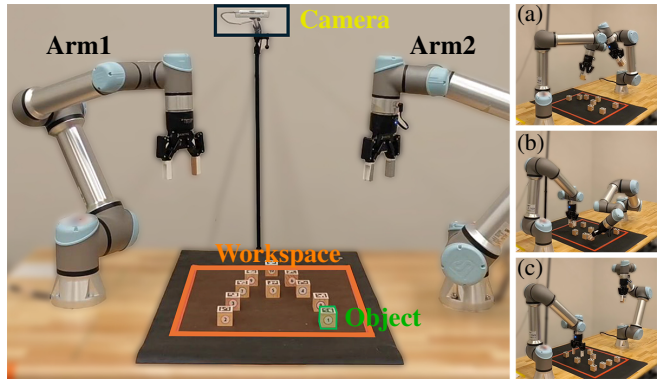

As we move from a single robot arm to dual-arm systems~\cite{chen2022cooperative, ZhangCASE22, shome2020synchronized, GaoYu22IROS} operating in close proximity, TAMP remains a key computation bottleneck. This is due to the dramatic increase in complexity from both task planning and motion planning. From the task planning side, the introduction of a second arm doesn't just allow the system to do two things at once, but can also do many new things, i.e., two arms can now collaboratively perform many tasks impossible for a single arm. This also means that the search space for possible task plans grows exponentially. 
From the motion planning side, the degree-of-freedoms (DoFs) of the system doubles from 6-7 for a single arm to 12+, making motion planning again challenging. While planners like Curobo~\cite{curobo} can compute nice motions for dual-arm systems for certain pre-specified start/goal configurations, as we have found in our project, doing so reliably for random start/goal configurations remains difficult.
Integrating task and motion planning further compounds the dual-arm TAMP challenge. 

In this paper, toward the goal of near-optimally solving task and motion planning challenges for dual-arm systems in real-time, we examine the long-horizon tabletop rearrangement task where many objects concentrated on a small tabletop must be rearranged using a dual-arm system (Fig.~\ref{fig:example}). 
Our framework, \emph{Synchronous Dual-Arm Rearrangement Planner} (\ours), follows the typical hierarchical algorithmic structure of TAMP solvers, bringing the following key contributions to dual-arm TAMP:
\begin{itemize}[leftmargin=4mm]
    \item \textbf{Dependency-Driven Dual-Arm Task Planning}. \ours's task planner (\ourst) efficiently resolves task dependencies through a layered approach, generating sub-tasks by ``peeling off'' tasks with increasing complexity. Unlike typical task planners, \ourst outputs multiple high-quality next-step sub-tasks. Taking full advantage of dual-arm systems, \ourst reduces the total sub-task counts as compared to optimal single-arm task planners~\cite{GaoYu22IROS}.
    \item \textbf{Sampling-Based Motion Generation and Optimization}. Borrowing the tried-and-true practice from sampling-based motion planning, \ours's motion planner (\oursm) examines multiple sub-tasks and for each sub-task, multiple potential grasp poses to select the highest quality (short-term) motion plan. This results in \oursm having a much higher success rate than~\cite{curobo}, which is a building block of \oursm.
    \item \textbf{Robust Failure Recovery}. In rare cases, \oursm may fail to plan an optimized trajectory using synchronous dual-arm motions, e.g., due to arm-arm collisions. When this happens, \ours detects it and engages a multi-stage fallback plan that gradually sacrifices solution quality to boost planning success rates. 
\end{itemize}

Fusing \ourst and \oursm, \ours delivers a highly effective, \emph{dual-arm native} solution for the tabletop rearrangement task. Extensive evaluation shows that \ours can handle a variety of challenging, dense tabletop rearrangement problems with $100\%$ success rates. In contrast, a baseline TAMP planner~\cite{GaoYu22IROS} only achieves a success rate of $85\%$. Simultaneously, \ours drastically shortens the task execution time by nearly three times, as compared with the baseline. Moreover, \ours is highly computationally efficient, taking only about $~5$ seconds to compute the task and motion for a single (dual-arm) sub-task, making it near real-time.

\section{Related Work}\label{sec:related}
\textbf{Task Planning and Multi-Object Rearrangement}.  
Task planning is a central problem in robotics, defining how symbolic actions are structured to achieve complex goals. 
Prior work spans diverse domains, from conflict resolution in multi-agent path finding \cite{LiIJCAI19} to everyday tasks such as table setting with commonsense knowledge from large language models (LLMs) \cite{ding2023task}. 
Within this scope, object rearrangement has drawn particular attention for both its practical significance and theoretical difficulty. 
Chang et al. \cite{chang2024lgmcts} proposed a language-guided MCTS framework for natural-language-driven tabletop rearrangement. 
Other studies addressed additional complexity, including varying object shapes, weights, or multi-layer stacking \cite{gao2023effectively, gao2023orla, xu2023optimal}, which further complicates task planning. 
From a computational standpoint, minimizing pick-and-place operations \cite{han2018complexity} or the number of temporarily displaced objects \cite{GaoFenYu21RSS} has been proven NP-hard.  

\textbf{Motion Planning}.  
Motion planning originated with 2D path-finding for point or polygonal robots. 
Early exact methods included cell decomposition \cite{la2011motion}, roadmap construction \cite{lozano1979algorithm}, and rotation-stacked visibility graphs \cite{ZhaYeYu25ICRA}, applicable mainly to $\real^2$ or $SE(2)$. 
To address higher-dimensional spaces, sampling-based algorithms such as RRT \cite{la2011motion}, and PRM \cite{bohlin2000path}, along with optimal variants \cite{karaman2011sampling, strub2020adaptively, gammell2020batch}, became standard, though they face exponential complexity growth in the number of DoFs.  
Gradient-based methods like TrajOPT \cite{schulman2013finding, schulman2014motion} and CHOMP \cite{zucker2013chomp} directly optimize trajectories, and recent semi-infinite programming formulations guarantee provably collision-free paths \cite{zhang2023provably, liang2024second}. 
Yet global optimality is not guaranteed, motivating hybrid approaches such as the graph of convex sets \cite{marcucci2023motion} and cuRobo \cite{curobo}, which leverage GPU parallelization for roadmap construction and trajectory optimization.  

Multi-arm coordination has gained interest for enabling parallel and flexible task execution \cite{smith2012dual, koga1994multi}. 
A straightforward extension applies planning methods in the full joint space, though dimensionality becomes prohibitive. 
Alternatives include hierarchical roadmaps merged into a super-graph \cite{gharbi2009roadmap}, conflict-based search (CBS) adaptations \cite{shaoul2024accelerating, shaoul2024unconstraining}, and shortcutting techniques to improve trajectory quality \cite{huang2025benchmarking}. 
Learning-based methods also show promise: Ha et al. \cite{ha2020learning} trained reinforcement learning policies from sampling-based demonstrations, achieving improved scalability and faster planning times.  

\textbf{Multi-Arm Task and Motion Planning}.  
Multi-arm task and motion planning (MATAMP) inherits the combinatorial difficulty of task planning and the geometric complexity of motion planning, yet it is crucial for real-world applications. 
Representative domains include cooperative assembly \cite{chen2022cooperative, huang2025apex}, roof bolting \cite{zhang2023multi}, tabletop rearrangement \cite{GaoYu22IROS, gao2024toward}, and bi-manual object retrieval from clutter \cite{wang2025learning, ahn2021coordination}. 
Classical approaches extend TAMP frameworks to handle inter-arm coordination and collisions: Chen et al. \cite{chen2022cooperative} coupled MILP-based task allocation with multi-agent motion planning for assembly, while Zhang et al. \cite{zhang2023multi} introduced a graph-guided MCTS for collaborative manipulation.  

Tabletop rearrangement has been a major focus, as it captures core MATAMP challenges. 
Shome et al. \cite{shome2021fast} optimized synchronous dual-arm rearrangement under monotone assumptions via MILP, while Gao et al. \cite{GaoYu22IROS, gao2024toward} developed dependency-graph and buffer-based methods for non-monotone cases. 
Learning approaches are also emerging: attention-based imitation learning improves robustness in dual-arm manipulation \cite{kimtransformer}, and DG-MAP \cite{parimi2025diffusion} leverages diffusion models to iteratively resolve collisions, scaling to as many as eight arms in simulation.

\section{Preliminaries}\label{sec:problem}
\subsection{Multi-Object Tabletop Rearrangement}
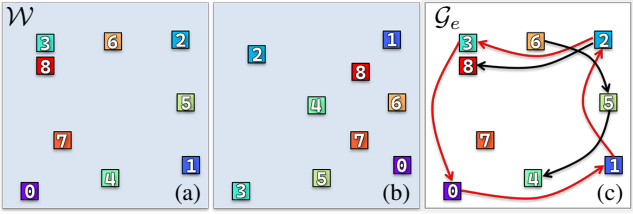
\begin{figure}[t]
\centering
\begin{overpic}[width=\linewidth]{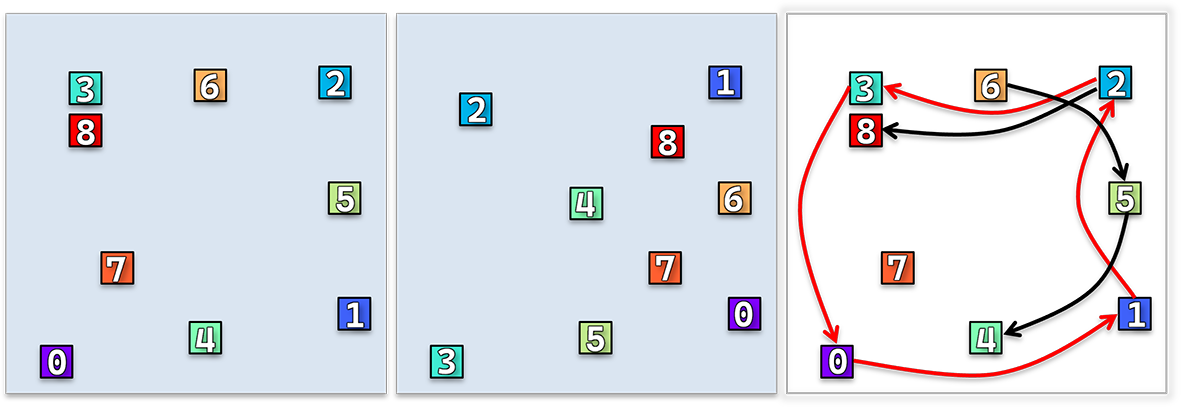}
    \put(27,3){\small{(a)}} 
    \put(59,3){\small{(b)}} 
    \put(92,3){\small{(c)}} 

    \put(1,30){$\small{\mathcal W}$}
    \put(67,30){$\small{\mathcal G_e}$}
\end{overpic}
\caption{Illustration of a tabletop rearrangement instance that will be used as a running example. $\mathcal W$ is the workspace (a) Start configuration $O^s$. (b) Goal configuration $O^g$. (c) The induced dependency graph $\mathcal G_e$. }
\label{fig:running_example}
\end{figure}

In a tabletop rearrangement problem, we work with a rectangular workspace $\mathcal W \subset \mathbb R^2$. There are two robot arms $r_1, r_2$, the end effector of each of which can reach the entirety of $\mathcal W$.  There are $n$ objects fully residing on $\mathcal W$, the tabletop, where object $i$ configuration is specified by $o_i = (x_i, y_i, \theta_i) \in SE(2)$ with $(x_i, y_i) \in \mathcal W$.\footnote{An object $i$ may be held by a robot arm $r_j$, in which case the object will have a \emph{special} configuration in $\{r_1, r_2\}$.} An \emph{arrangement} of the objects is specified as $O = \{o_1, \dots, o_n\}$. An arrangement $O$ is \emph{feasible} if there is no collision between any pair of objects $i$ and $j$, $i \neq j$, considering the objects' physical footprint. For example, Fig.~\ref{fig:running_example} (a)(b) show two feasible configurations of eight cuboids (top-down view). We define the dual-arm rearrangement problem as follows. 

\begin{problem}[$k$-Arm Tabletop Rearrangement]
Given two feasible arrangements $O^s = \{o_1^s, \dots, o_n^s\}$ and $O^g = \{o_1^g, \dots, o_n^g\}$, a $k$-arm tabletop rearrangement problem asks for a sequence of intermediate arrangements $\{O^s = O^1, \dots, O^m=O^g\}$ such that:
\begin{enumerate}[leftmargin=6mm]
    \item $O^i$, $1 \le i \le n$, is a feasible arrangement.
    \item For each consecutive pair of arrangements $(O^i, O^{i+1})$, $1 \le i < n$, there exists feasible motions for the robots to transition from $O^i$ to $O^{i+1}$ in which each arm manipulates at most a single object.
\end{enumerate}
\label{prob:datr}
\end{problem}

In this work, $k=2$. Fig.~\ref{fig:running_example} illustrates an instance of Problem~\ref{prob:datr} where Fig.~\ref{fig:running_example}(a) corresponds to $O^s$ and Fig.~\ref{fig:running_example}(b) corresponds to $O^g$. This instance will be used a running example for explaing how \ours work.

We note that Problem~\ref{prob:datr} is a more challenging version than a similar problem from~\cite{GaoYu22IROS} as a suction-based grasping model is assumed by~\cite{GaoYu22IROS} to grasp cylindrical objects, which is much easier than using $2$-finger grippers to grasp non-cylindrical objects from the sides. 

\subsection{Dependency Graph}
In rearranging objects, a key challenge is the untangling of object dependencies. To effectively accomplish this, we leverage a data structure called the dependency graph (DG)~\cite{han2018complexity}, induced by the start and goal configurations of an object rearrangement problem. In a DG $\mathcal{G} = (V, E)$, vertex $v_i \in V$ corresponds to object $i$, and a directed edge $(v_i \to v_j) \in E$ indicates that object $i$ cannot be placed at its goal pose until object $j$ has been moved away. 

Fig.~\ref{fig:running_example}(c) provides the DG induced by Fig.~\ref{fig:running_example}(a),(b) as the start and goal configurations, respectively. In the DG $\mathcal G_e$, object 7 has no dependencies, suggesting it can be directly ``solved''. Similarly, objects 4 and 8 do not depend on other objects (no outgoing edge) and can be directly moved to their goals. All other objects, on the other hand, have outgoing edges, meaning that they can not be solved without moving some other objects first. In particular, the red edges form a \emph{directed cycle}, meaning that they form a cyclic dependency among themselves. Resolving such cyclic dependencies requires either picking up and holding an object or relocating an object to a temporary location in $\mathcal W$.

\section{Synchronous Dual-Arm Rearrangement}\label{sec:algorithm}
Using the example illustrated in Fig.~\ref{fig:running_example}, in this section, we elaborate on how our Synchronous Dual-Arm Rearrangement Planner (\ours) framework functions. 
\vspace{-2mm}
\subsection{Dependency-Driven Dual-Arm Task Planning}
\subsubsection{Structure of the Dependency Graph}
A DG is constructed in a straightforward manner. Given two configuration $O^s$ and $O^g$, geometric conflicts are examined for each $o_i^g$ and $o_j^s$; an edge $(v_i \to v_j)$ is added if $o_i^g$ and $o_j^s$ overlap spatially. Collecting all these dependency edges then yields a DG for ($O^s, O^g$). 

The task planner of \ours, \ourst, identifies from a DG the following: 
\begin{itemize}[leftmargin=4mm]
    \item \textbf{Independent Objects:} These are vertices with zero out-degree and \jh{I added an 
"and"} can be moved immediately, e.g., objects 4, 7, and 8 in Fig.~\ref{fig:running_example}.
    \item \textbf{Chains:} Chains are directed paths in a DG that impose strict order with which objects can be moved sequentially, e.g., objects 4, 5, and 6 in Fig.~\ref{fig:running_example}. 
    \item \textbf{Cycles:} Cycles are strongly connected components that require more coordination to resolve, e.g., objects 0, 1, 2, and 3 in Fig.~\ref{fig:running_example}.
\end{itemize}
In Fig.~\ref{fig:running_example}, there are nine objects to be rearranged (labeled 0--8), forming a dependency graph (DG) $\mathcal G_e$ that includes one (directed) cycle (objects 0--3), one chain (objects 4--6), and two independent objects (objects 7 and 8).

\subsubsection{Sub-Task Decomposition and Assignment} Similar to~\cite{GaoYu22IROS}, \ourst derive possible task plans through a DG-driven analysis, decomposing the rearrangement problem into tractable sub-tasks and generating a task plan that respects all constraints. However, the task planner in~\cite{GaoYu22IROS} first searches for an optimal plan for \jh{would "each single arm" better than "a single arm"?} a \emph{single} robot arm, which does not effectively leverage a dual-arm system's capabilities. \ourst, instead, seeks to break down a problem directly using two robot arms, via simultaneous \emph{node removal sequencing} and \emph{node-arm assignment}. As will be demonstrated in the evaluation, compared with~\cite{GaoYu22IROS}, \ourst generally takes fewer/same number of actions.

\textbf{Node Removal Sequencing}. \ourst begins by rearranging all independent objects, which is straightforward. Doing so effectively \emph{deletes} all \emph{orphan nodes} on the DG. For $\mathcal G_e$ in Fig.~\ref{fig:running_example}, object $7$ gets removed. \jh{Then, \ourst looks at chains, resolving their \emph{leaf nodes} one by one since they have no dependencies. In $\mathcal G_e$ , the initial leaf nodes are 4 and 8. After removing 4, node 5 becomes a new leaf node, and so on.} Then, \ourst looks at chains, the \emph{leaf nodes} of which can be resolved one by one because they have no dependencies. In $\mathcal G_e$, these are $4$ and $8$ initially. After $4$ is moved and deleted from $\mathcal G_e$, $5$ becomes a leaf node, and so on. 
After removing all orphan and leaf nodes, the DG is either empty or consists only of cycles. In the case of $\mathcal G_e$, only the cycle $0 \to 1 \to 2 \to 3 \to 0$ remains. These remaining cycles are handled last by \emph{cycle breaking}, \jh{not sure what we want to express here}which may create new orphan and leaf nodes. \ourst distinguishes between two cases for cycles. For a $2$-object cycle, a direct dual-arm swap is sufficient. For longer cycles, \ourst selects two consecutive objects (e.g., $0$ an $1$ in $\mathcal G_e$) such that one is temporarily relocated to a \emph{buffer} because it cannot be directly placed at its goal (if $0$ and $1$ are selected for $\mathcal G_e$, $1$ must be placed at a buffer because of the $1\to 2$ dependency). \jh{\ourst merely indicates the need for a buffer, while the motion planner (\oursm) determines its actual placement.} \ourst only marks that a buffer is needed, leaving the motion planner (\oursm) to find it. 

For $\mathcal G_e$, one possible sequence is $7, 8, 4, 5, 6, 0, 1, 3, 2, 1$, where $1$ appears twice because it needs to be temporarily relocated. Such sequence is not unique; \ourst always works with multiple sequences implicitly. 

\textbf{Node-Arm Assignment}. As the DG is being decomposed, removed/updated objects are \emph{assigned} to arms. Essentially, this is done by taking pairs of nodes in the front of node removal sequences and \jh{Need some clearify. Does this means: in the pair, the first object assigned to left arm, and second object assigned to the right? Is the position of item considered? (i.e. the item closer to the left assigned to the left arm)} pairing them up. For $\mathcal G_e$, the first assignment pair can be $(4, 7)$, $(7, 8)$, $(4, 8)$, and so on, where the order is interchangeable (e.g., $(4,7)$ $\equiv$ $(7,4)$, likewise for $(7,8)$ and $(4,8)$).
\vspace{-0.3mm}
\subsubsection{Algorithm Sketch}
We now outline \ourst. Let $r_1$ and $r_2$ denote the two robot arms. A task plan is denoted by $(i, j)$, where $i \ne j$: arm $r_1$ rearranges $(s_i, g_i)$ and arm $r_2$ rearranges $(s_j, g_j)$. $(s_i, g_i)$ are the \emph{current} start and goal configurations for $i$, respective, which may be different from $(o_i^s, o_i^g)$. The same goes for $j$. Each arm maintains the following state information: (1) Gripper state, which may be OPEN or CLOSE, (2) Assigned object ID, and (3) Grasp angle, (4) Execution stage $ToStart$ and $ToGoal$, indicating whether $r_1$ and $r_2$ are heading to start poses. A sketch of \ourst is outlined in Alg.~\ref{tp}.
\vspace{-1mm}

\begin{algorithm}
\begin{small}
\vspace{0.025in}
\textbf{Input:} Start Configuration $S$, Goal Configuration $G$ \\
\textbf{Output:} Task plan for the next step
\vspace{0.025in}

Build the dependency graph $\text{DG}(S, G)$

\If{Both arms are $ToGoal$}{
    Set gripper actions to OPEN\\
    Maintain current assignments and angles \\
    \Return task plan
}

\If{Both arms are $ToStart$}{
    taskList $\leftarrow \varnothing$\\
    \uIf{only one object left}{
        Assign one arm to complete it, while the other goes back to the retract position \\
        \Return task plan
    }
    \uElseIf{multiple independent objects}{
        Append the permutations of independent objects to taskList
    }
    \uElseIf{a chain exists in DG}{
        Append the terminal pair to taskList
    }
    \ElseIf{a cycle exists in DG}{
        Append all adjacent pairs to taskList.\\
        Set \textit{NeedBuffer} flag to true
    }
    Set both gripper actions to CLOSE, assign taskList\\
    \Return task plan
}
\vspace{0.025in}
\caption{\ourst Task Planner}\label{tp}
\end{small}
\end{algorithm}
\vspace{-1mm}

The task plan structure encapsulates all relevant execution parameters, including: (1) Potential task list with the form $\{(i^{(1)}, j^{(1)}), (i^{(2)}, j^{(2)}), \ldots\}$, (2) Gripper actions for $r_1$ and $r_2$ that can be OPEN or CLOSE, which will be executed after reaching the start/goal of the arm, (3) Assigned object IDs and grasp angles for both arms, and (4) Buffer needed flag.

\subsection{Sampling-Based Synchronous Motion Generation}
\oursm, our motion planner, leverages Nvidia tools, assembled within cuRobo~\cite{curobo}, to perform GPU-SIMD-based accelerated motion generation and optimization. In this study, motions for the two robot arms are synchronized at the sub-task level to limit the search space that is explored, because \ourst already generates many potential sub-tasks for every pair of robot motions and \oursm further explores many possible parameters to select the best sub-task. 

\subsubsection{Sub-Task Selection and Instantiation} The general idea behind our motion planner, \oursm, is to sample many possible sub-tasks and select the one with the lowest cost. In other words, \oursm takes a best-first approach in arm motion generation. The amount of sampling done is determined by the amount of computation budget, the amount of parallelism, and the feasibility of individual samples. 

\begin{figure}[h!]
\centering
\includegraphics[width=\linewidth]{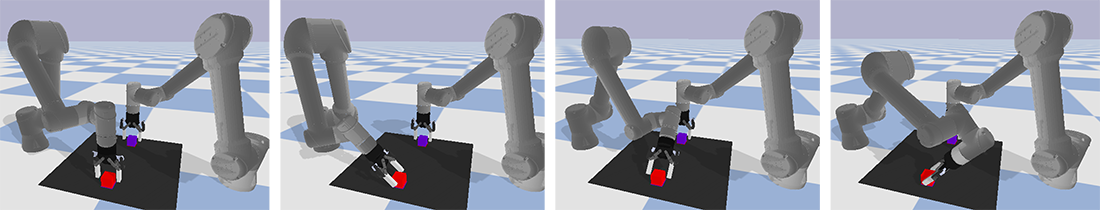}
\caption{Illustration of possible sampled grasp poses for grasping the two cuboids. For the arm on the right, the grasp poses are the same as the top-down ones. For the left arm, each is a different pose from a different approaching angle.}
\vspace{-2mm}
\label{fig:grasp_angles}
\end{figure}

In a nutshell, the ``best'' sub-task is selected based on inverse kinematics (IK) feasibility. For each sub-task in the current sub-task list as generated by \ourst, recall that it is in the form of $(s_i, g_i)$ for arm $r_1$ and $(s_j, g_j)$ for arm $r_2$. For generate arm motion, object poses $(s_i, g_i)$ and $(s_j, g_j)$ must be paired with corresponding arm poses. $s_i$ and $s_j$ are already bound to the ending pose of the previous arm motion; \oursm needs to sample potential arm poses for $g_i$ and $g_j$. 
To do this, multiple 2-finger gripper approaching angles are examined in an expanding manner (shown in Fig.~\ref {fig:grasp_angles}). \dz{Added the grasp angles in simulation here.} That is, top-down grasps (for a cuboid, there are two of these) will be tried first, after which sideway grasps with increasingly larger approaching angles will be attempted. The process also instantiates the sub-task for the robot arms. 

\begin{figure*}[t]
\centering
\begin{subfigure}[b]{0.48\textwidth}
\frame{
\includegraphics[width=\linewidth]{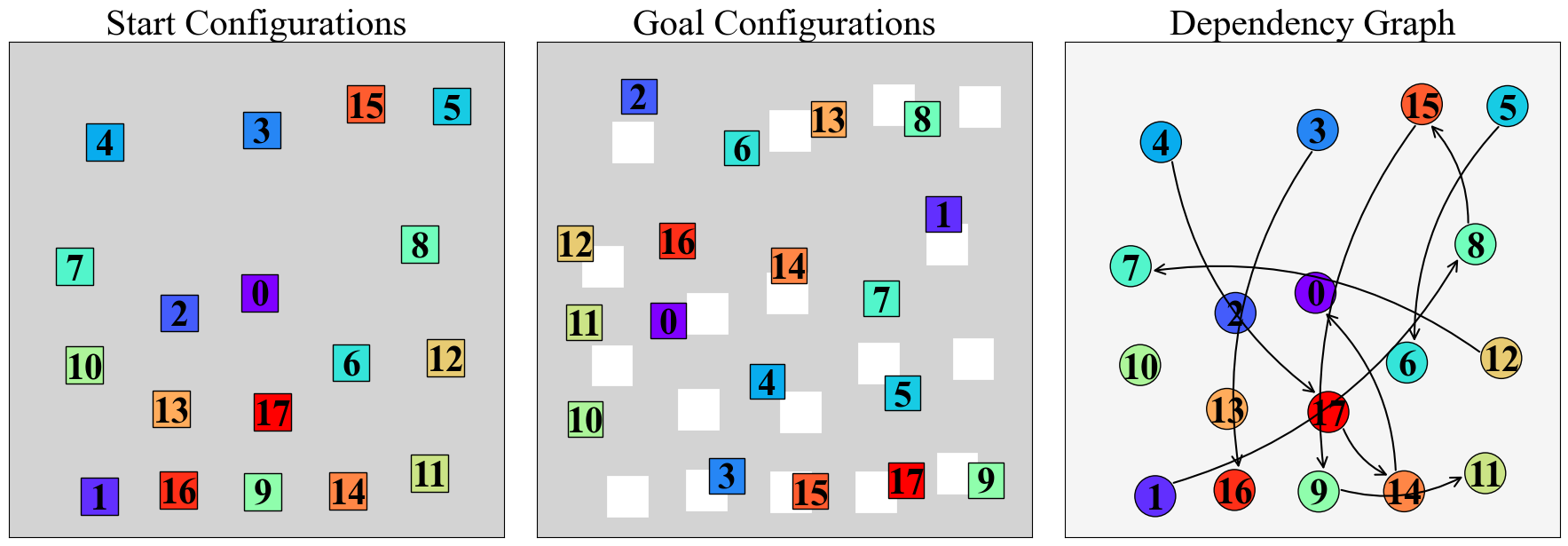}}
\end{subfigure}
\quad
\begin{subfigure}[b]{0.48\textwidth}
\frame{
\includegraphics[width=\linewidth]{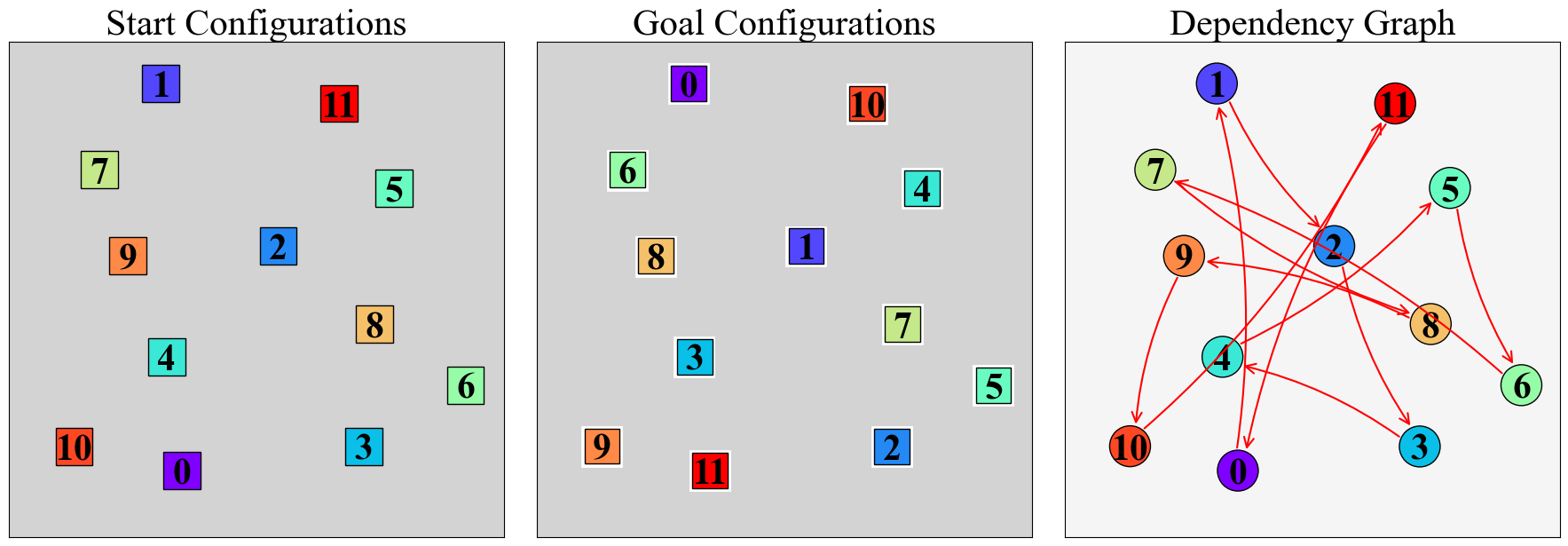}}
\end{subfigure}\\
\vspace{1mm}
\begin{subfigure}[b]{0.48\textwidth}
\frame{
\includegraphics[width=\linewidth]{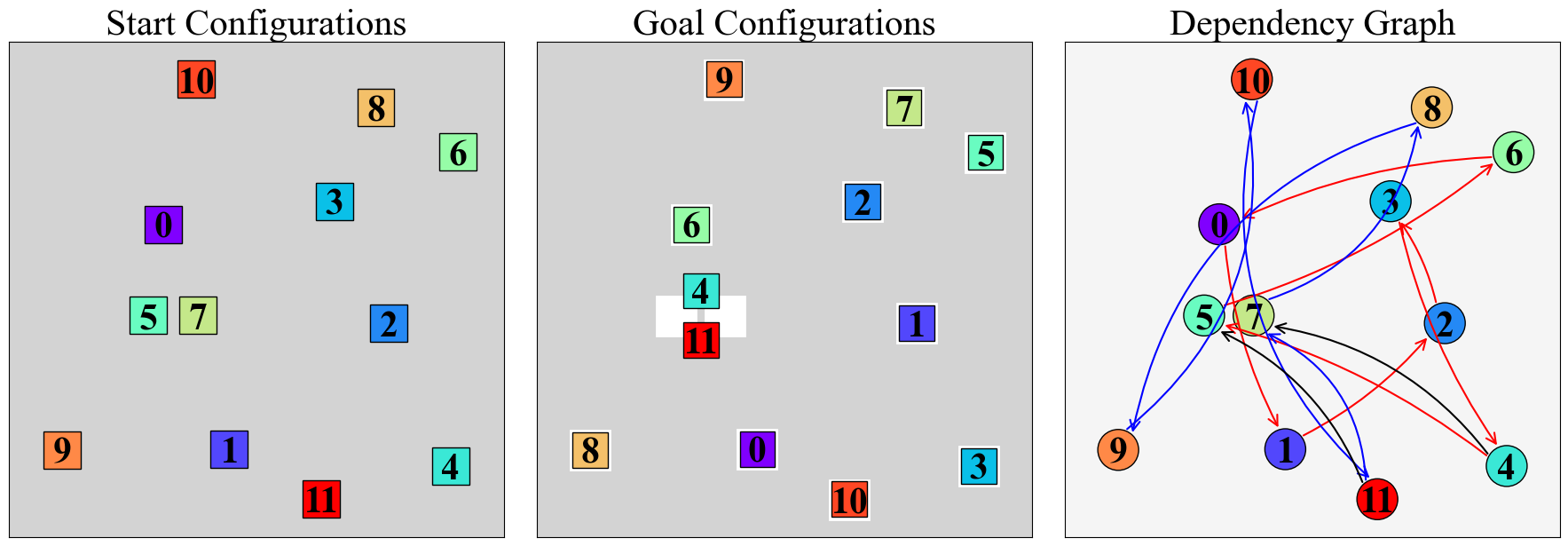}}
\end{subfigure}
\quad
\begin{subfigure}[b]{0.48\textwidth}
\frame{
\includegraphics[width=\linewidth]{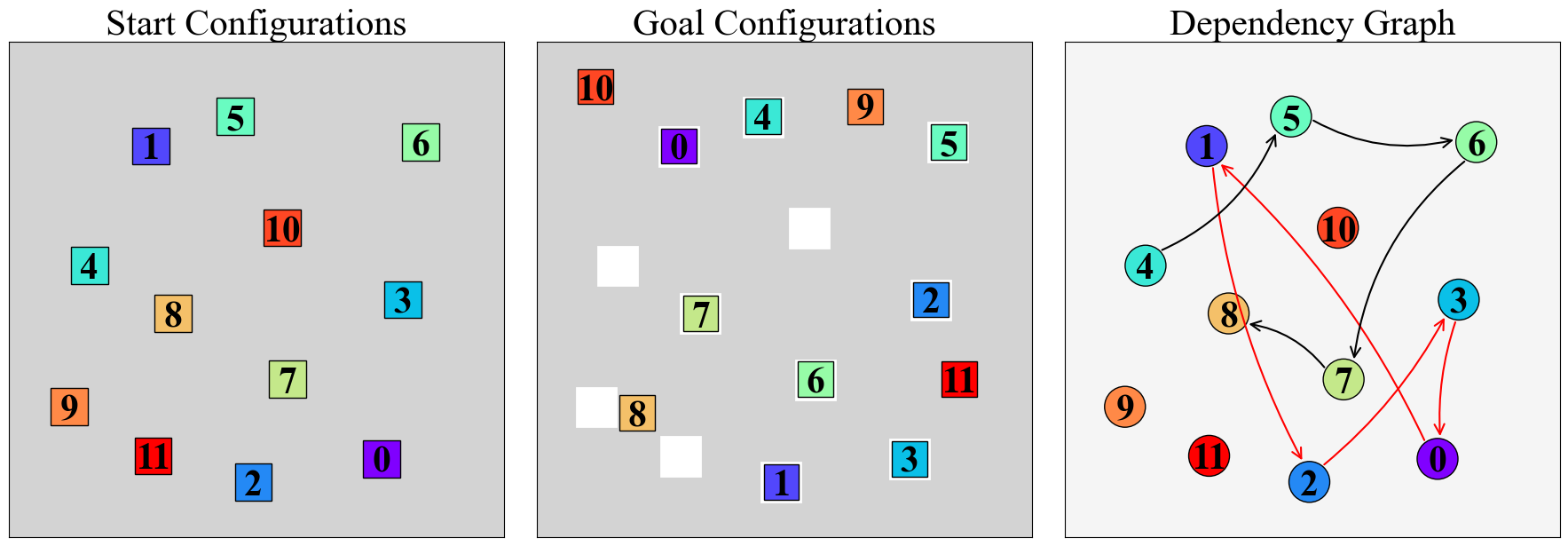}}
\end{subfigure}
\caption{Examples of start and goal configurations along with their corresponding dependency graphs. In the goal panels, the start configurations are also shown as white squares for reference.  
The four categories are shown: \textbf{(top left)} random configuration, 
\textbf{(top right)} single-cycle configuration, 
\textbf{(bottom left)} double-cycle configuration, and 
\textbf{(bottom right)} mixed configuration containing independent tasks, chains, and cycles. 
Black arrows denote chain dependencies in the task graph, while red and blue arrows represent different cycles.
}
\label{fig:testcases}
\end{figure*}
When a sub-task indicates that a temporary buffer location is needed, \oursm will sample $k$ non-overlapping positions in the workspace as potential buffers. Grasp poses for these will be generated similarly. 


For actual IK computation, we employ the SIMD-based batch IK computation capability of~\cite{curobo}. In a single batch, on an Nvidia RTX 4090 with 24GB RAM, it is possible to compute $\sim$256 IKs. On average, for each sub-task selection, 20 batches are sufficient for sifting through enough pose candidates to select a high-quality sub-task with $\sim$1.7s to compute all the IKs.  
\jy{Doesn't this 1.7 seconds conflict with the 0.52 seconds mentioned later?}
\dz{We don't need to solve batch IK for every step, only when a buffer is needed. To check the IK feasibility, we only need one, and that is so much faster. Also, when the arms have already grasped the objects, there's no need to solve IK at all because it's already solved when the start and goal pair are selected.}

\begin{algorithm}
\begin{small}
\vspace{0.025in}
\textbf{Input:} Potential task list $\mathcal{T}$ \\
\textbf{Output:} Executable motion plan
\vspace{0.025in}

$(i, j) \leftarrow$ \textsc{SelectBestTask}$(\mathcal{T})$ \\
$(s_i, s_j) \leftarrow$ current arm poses\\
$(t_i, t_j) \leftarrow$ target poses for objects $i$ and $j$ depending on execution stages \\
$(\alpha_1, \alpha_2) \leftarrow$ selected grasp angles

$P \leftarrow$ \textsc{cuRobo.Plan}$(s_i, t_i, s_j, t_j, \alpha_1, \alpha_2)$ \\
\If{$P$ is valid}{ \Return $P$ }

$P_1\leftarrow$ \textsc{ArmUntangle}$(s_i, t_i, s_j, t_j, \alpha_i, \alpha_j)$ \\
Get new arm poses $s_i', s_j'$ from $P_1$\\
$P_2 \leftarrow$ \textsc{cuRobo.Plan}$(s_i', g_i, s_j', g_j, \alpha_1, \alpha_2)$ \\
\If{$P_1$ and $P_2$ are valid}{ \Return $P_1 + P_2$ }
$P_1^{single} \leftarrow$ \textsc{cuRobo.Plan}$(s_i, \text{retract}_1, s_j, t_j, \alpha_1, \alpha_2)$ \\
$P_2^{single} \leftarrow$ \textsc{cuRobo.Plan}$(\text{retract}_1, t_i, t_j,\text{retract}_2, \alpha_1, \alpha_2)$ \\
\If{$P_1^{single}$ and $P_2^{single}$ are valid}{ \Return $P_1^{single} + P_2^{single}$ }
\Return \texttt{failure}
\vspace{0.025in}
\caption{\textsc{\oursm Motion Planner}} \label{alg:dual-arm-plan}
\end{small}
\end{algorithm}
\subsubsection{Motion Planning with Arm Untangling}
After a sub-task is selected and instantiated for the robot arms, motion plans can be generated. \oursm uses a two-stage process for completing this process. First, \oursm attempts to generate smooth optimized motions from the current arm positions to the target positions (can be start or the goal of the arm depending on the Execution Stages) using the \textsc{MotionGen} module from cuRobo~\cite{curobo}. Despite carefully selecting poses for the robots, \textsc{MotionGen} can fail, in which case \oursm falls back to a rule-based planner, \textsc{ArmUntangling}, that follows a set of rules to untangle the motions of the two robot arms to increase the motion planning success rates. 

In rare cases, \textsc{ArmUntangling} may also fail, in which case \oursm falls back to sequential planning and execution. Outline of \oursm is outlined in Alg.~\ref{alg:dual-arm-plan}. Our overall method, \ours, calls \ourst and \oursm sequentially until an instance is fully resolved. 

\section{Evaluation}\label{sec:evaluation}
We evaluate \ours through a series of simulated experiments designed to assess its efficiency, robustness, and scalability across diverse task settings. 
All algorithms are implemented in \texttt{Python} and executed on a workstation equipped with an Intel Core i9-14900K CPU and an NVIDIA RTX 4090 GPU.  

Four categories of test cases are generated with distinct dependency graph structures:  
\begin{enumerate}[leftmargin=5mm]
    \item \textbf{Random (R):} random start/goal configurations.  
    \item \textbf{Single cycle (S):} configurations inducing one DG cycle.  
    \item \textbf{Double cycle (D):} configurations inducing two distinct DG cycles.  
    \item \textbf{Mixed (M):} configurations inducing a mixture of independent objects, chains, and cycles.  
\end{enumerate}

For each category, randomness is injected whenever appropriate in creating the start/goal configurations. Representative cases of the test cases are illustrated in Fig.~\ref{fig:testcases}.
We denote each case as \emph{R\#}, \emph{S\#}, or \emph{D\#}, where the number indicates the number of objects, and as \emph{M\#} for mixed cases, where the number is simply an index as all mixed cases use 12 objects.

We evaluate \ours against the previous state-of-the-art~ \cite{GaoYu22IROS} as the baseline. Beyond the direct comparison, we also consider several hybrid configurations to independently evaluate \ourst and \oursm (TP means task planner and MP means motion planner): (1) Baseline TP $+$ \oursm, (2) \ourst$+$ Baseline MP,
(3) \ourst$+$ cuRobo, and (4) Baseline TP $+$ cuRobo.


\subsection{Task Planning Performance}
We first evaluate the quality of the task plans produced by \ourst in comparison to the baseline. 
The number of actions in a plan is computed by counting each task assigned to either arm as one action and summing across both arms. 
Although \ourst may generate multiple potential tasks at each step, the motion planner ultimately selects one option, and this determined sequence is used for evaluation.  

\begin{figure*}[t]
    \centering
    \includegraphics[width=\textwidth]{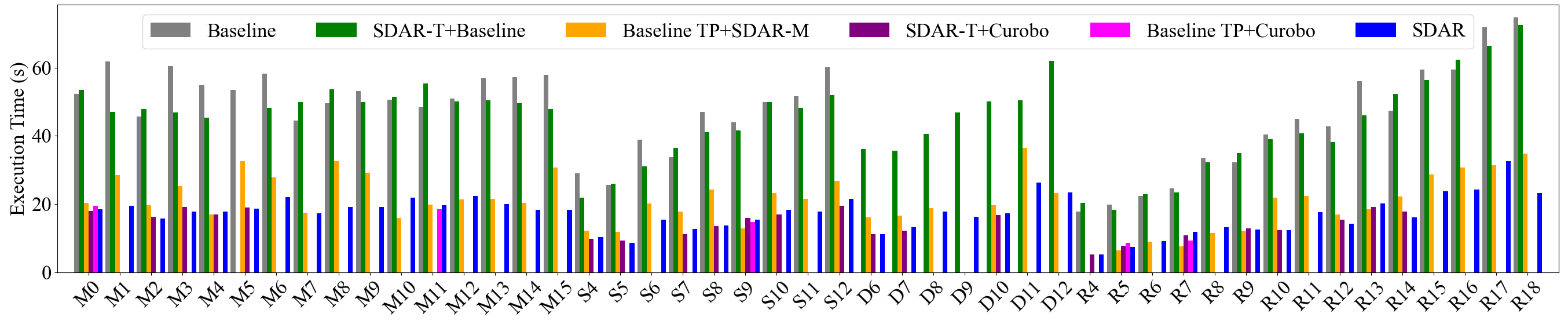}
    \caption{Execution time for all simulation-evaluated test cases. Where the horizontal legends are as defined at the start of this section (e.g., R10 means the random setting with 10 objects). Missing bars for a given test case and algorithm combination indicate the method failed to produce a valid solution.}
    \label{fig:execution_time_comparison}
\end{figure*}

\begin{figure}[h]
    \centering
    \includegraphics[width=\linewidth]{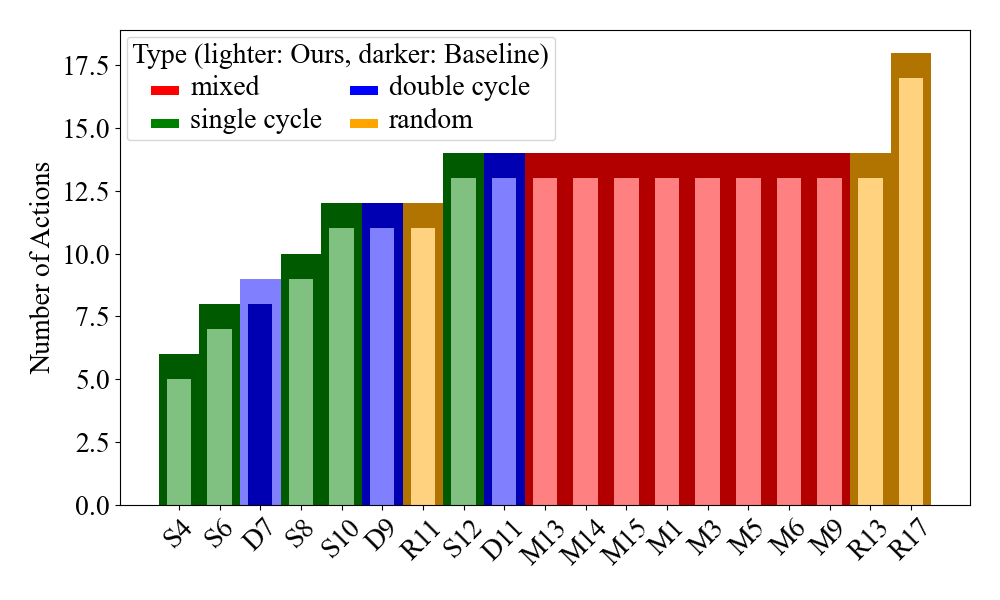}
    \caption{Absolute number of actions per test case for instances where our method and the baseline differ. Lighter bars correspond to our method and darker bars to the baseline, with colors indicating configuration type (red = mixed, green = single cycle, blue = double cycle, yellow = random).}
    \label{fig:num_actions_barplot}
\end{figure}

Fig.~\ref{fig:num_actions_barplot} shows the absolute number of actions for the subset of test cases where \ourst and the baseline produce different results. 
Lighter bars correspond to \ourst and darker bars to the baseline, with colors indicating the four dependency graph categories (red/M, green/S, blue/D, yellow/R).  
In nearly all cases, \ourst uses fewer or an equal number of actions.  
An exception occurs in D7 (double cycle, 7 objects), where the baseline produces a plan with fewer actions by holding one object continuously. The baseline plan, however, is very inefficient because that arm is essentially idling when holding an object, leaving the other arm to do all the work.  

\begin{figure}[h]
    \centering
    \includegraphics[width=\linewidth]{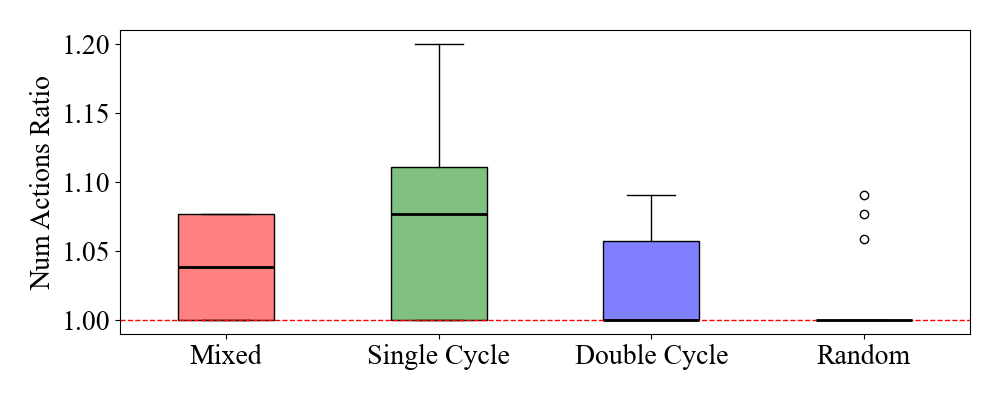}
    \caption{Ratio of the number of actions (baseline/\ours) grouped by dependency type without case D7.}
    \label{fig:num_actions_ratio_boxplot}
\end{figure}

Fig.~\ref{fig:num_actions_ratio_boxplot} reports the ratio of action counts of baseline over \ourst.
Values above $1.0$ indicate that the baseline required more actions. 
\ourst consistently produces shorter or equal plans in mixed, single-cycle, and double-cycle settings, while achieving parity in random configurations.  
These results demonstrate that our method generally reduces the number of actions compared to the baseline, which builds on an optimal single-arm task planner~\cite{GaoFenHuaYu23IJRR}. 

\begin{figure}[h!]
    \centering
    \includegraphics[width=\linewidth]{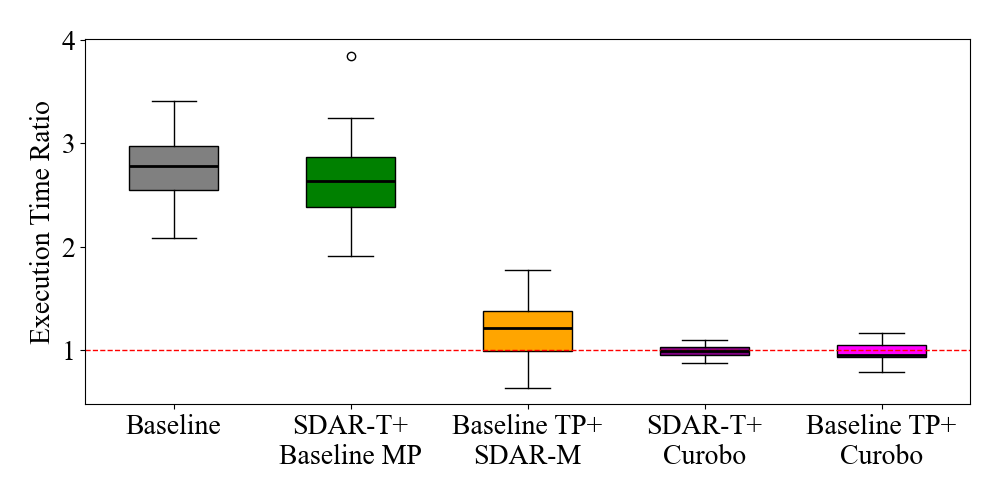}
    \caption{Execution time ratios compared to our method (red dashed line = $1.0$). Motion planner quality is the dominant factor for execution efficiency.}
    \label{fig:execution_time_ratios_boxplot}
\end{figure}

\begin{figure}[h]
    \centering
    \includegraphics[width=\linewidth]{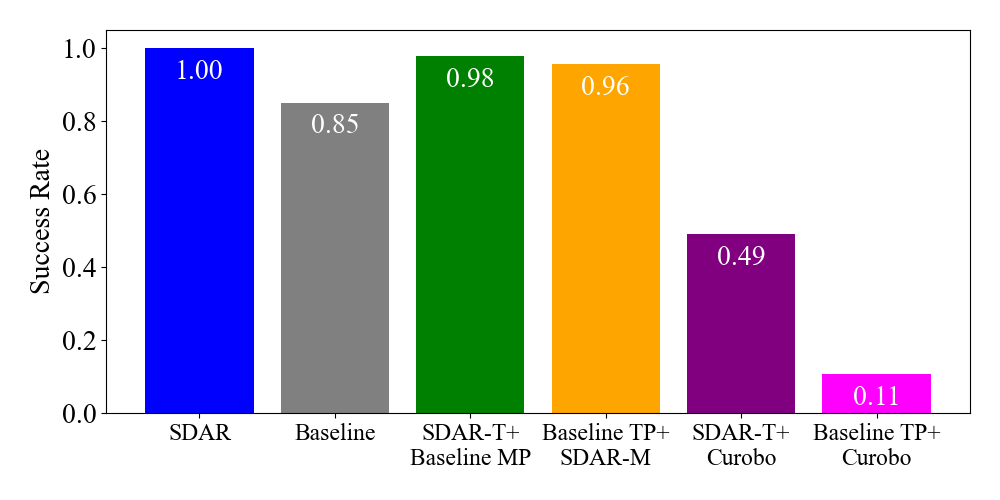}
    \caption{Success rates achieved by different task planner and motion planner combinations over all test cases.}
    \label{fig:success_rate_comparison}
\end{figure}

We now evaluate the end-to-end performance of \ours.
Fig.~\ref{fig:success_rate_comparison} reports the success rates, defined as the ability to generate a feasible dual-arm plan that can be executed without collision or deadlock.  
Our full method achieves a $100\%$ success rate, demonstrating the robustness of \ours's multi-level search architecture.
The baseline method achieves $85\%$ success, while hybrid variants that mix task and motion planners perform better but still fall short of \ours. 
Baseline TP$+$cuRobo in particular suffers from extremely low success ($11\%$), showing that even strong motion planning cannot compensate for weak task planning. 
Similarly, \ourst$+$cuRobo achieves only $49\%$ success, underscoring the importance of robust motion planning strategies.
\vspace{-2mm}
\subsection{Overall Task and Motion Planning Performance}
\begin{figure*}[t]
    \centering
    \includegraphics[width=\textwidth]{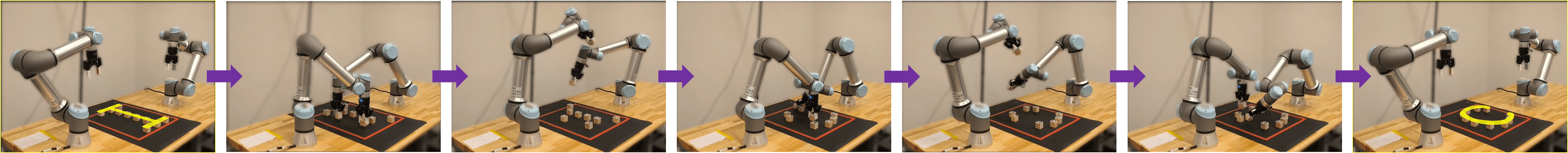}
    \caption{Snapshots from the real-robot experiment on dual-arm rearrangement. 
    Two UR5e manipulators with Robotiq grippers rearrange ArUco-tagged cubes from the letter ``I'' to ``C'' as part of forming ``ICRA2026''.}
    \label{fig:real_robot_icra}
\end{figure*}
Fig.~\ref{fig:execution_time_ratios_boxplot} summarizes execution time performance across categories relative to our method, \ours.
The results suggest execution efficiency is largely dictated by the motion planner. 
When paired with the same motion planner (either ours or the baseline’s), our task planner (\ourst) consistently leads to lower execution time compared to the baseline task planner, delivering over $60\%$ execution time savings. 
At the same time, replacing a weaker motion planner with a stronger one yields the largest gains regardless of the task planner.  
This explains why both \ourst$+$cuRobo and Baseline TP$+$cuRobo achieve very low execution time ratios: cuRobo produces high-quality motion plans efficiently, but without fallback mechanisms, these methods fail frequently, as reflected in their success rates. 
Fig.~\ref{fig:execution_time_comparison} presents the absolute execution time across cases and methods for a quick assessment of all methods. 

Taken together, these results demonstrate that robust and efficient dual-arm planning requires not only high-performance task planner and motion planner individually, but also tight integration between the two, which underlies the design principle of \ours.
\subsection{Computation Time}
We profiled \ours across all test cases and observed that \oursm takes nearly all planning time. Within \oursm, $9\%$ of computation time (averaging 0.52 seconds per step) is spent on IK computation for plan selection, while $91\%$ (5.17 seconds per step) is used to compute motion plans for the selected task plan. This further corroborates the importance of an efficient motion planner with a high success rate. 

\subsection{Real-Robot Experiment}

We conducted real-world experiments using two UR5e manipulators equipped with Robotiq 2F-85 grippers to confirm that \ours indeed produces high-quality, collision-free plans. The task asks the two robots to rearrange 10 wooden cuboids to form letters in ``ICRA2026'', one by one. 
Our planner successfully generated collision-free trajectories for both arms, which were executed in parallel on robot hardware.  
Snapshots from the experiment are shown in Fig.~\ref{fig:real_robot_icra}, highlighting an intermediate step where the arrangement transitioned from the letter ``I'' to ``C''. 
A full video demonstration is also included, showing the complete execution.

\section{Conclusion and Discussions}\label{sec:conclusion}
This work introduced the Synchronous Dual-Arm Rearrangement Planner (\ours), a task and motion planning framework designed for solving long-horizon tabletop rearrangement tasks with complex (non-monotone) object dependencies. \ours tightly integrates dependency-driven task planning with robust synchronous dual-arm motion generation. By coupling layered decomposition of dependency graphs with sampling-based motion optimization and fallback strategies, \ours consistently delivers high-quality solutions, near real-time efficiency, and a $100\%$ success rate across challenging long-horizon rearrangement tasks. Comprehensive evaluation confirms that \ours delivers state-of-the-art task planning and motion planning performances separately, and as a whole, achieves $60$+$\%$ reduction in task execution time. Real-robot experiments with dual UR5e manipulators demonstrate the framework's ability to transfer readily and reliably to hardware. 

Looking ahead, many promising avenues remain; we mention a few here. 
From the tasking planning perspective, dual-arm systems provide the unique ``swapping'' primitive that is beyond the capability of a single arm. It is interesting to explore what a $k$-arm system can achieve with regard to task planning. 
From the motion planning side, currently, we are exploring improving motion planning efficiency through better parallelization and learning-guided sampling to further reduce latency toward true real-time dual-arm deployment.
Beyond improving task and/or motion planning performance, to make frameworks like \ours more useful for general-purpose manipulation, richer perception, and online feedback should be tightly integrated into the task-motion planning loop.
Lastly, getting back to \ours, an elephant in the room is the word ``synchronous''. Asynchronous dual-arm coordination will drastically expand the solution space and can potentially lead to a far superior TAMP solution. However, how to tame the ensuing combinatorial explosion?



{\small
\bibliographystyle{formatting/IEEEtran}
\bibliography{bib/jingjin}
}

\end{document}